\documentclass[pageno]{jpaper}
\usepackage[normalem]{ulem}
\usepackage{comment}
\usepackage{graphicx}
\usepackage{amsmath}
\usepackage{amssymb}
\usepackage{multirow}
\usepackage{multicol}
\usepackage{wrapfig}
\usepackage{cite}
\usepackage{pifont}
\usepackage{color}
\usepackage{flushend}

\newcommand{\squishlist}{
 \begin{list}{$\bullet$}{
  \setlength{\itemsep}{0pt}      
  \setlength{\parsep}{0pt}
  \setlength{\topsep}{3pt}       
  \setlength{\partopsep}{0pt}
  \setlength{\listparindent}{-2pt}
  \setlength{\itemindent}{-5pt}
  \setlength{\leftmargin}{0.5em} 
  \setlength{\labelwidth}{0em}
  \setlength{\labelsep}{0.5em} 
 } 
}

\newcommand{\squishend}{
 \end{list} 
}

\newcommand{\ignore}[1]{}  

\begin{document}

\twocolumn[ 	
	\begin{@twocolumnfalse}
    \title{WRPN \& Apprentice: \protect\\Methods for Training and Inference using Low-Precision Numerics}
    \author{Asit Mishra \& Debbie Marr}
    \date{Accelerator Architecture Lab, Intel Labs}
	\maketitle
	\end{@twocolumnfalse}
]
\thispagestyle{empty}

\begin{abstract}
Today's high-performance deep learning architectures involve large 
models with numerous parameters. 
Low-precision numerics
has emerged as a popular technique to
reduce both the compute and memory requirements of these large models.
However, lowering precision often leads to accuracy degradation.
We describe three software-based schemes whereby 
one can both train and do efficient inference using 
low-precision numerics without hurting accuracy.
Finally, we describe an efficient hardware accelerator that can take 
advantage of the proposed low-precision numerics.
\end{abstract}
\section{Introduction} \label{sec:intro}
\textbf{Background}: Using low-precision numerics is a 
promising approach to lower compute and memory 
requirements of convolutional deep-learning
workloads.
Operating in lower precision mode reduces computation as
well as data movement and storage requirements.
Due to such efficiency benefits, there are many existing works 
which propose
low-precision deep neural networks (DNNs)~\cite{NN-FewMult, 
LogNN, StochasticRounding, GoogleLowPrecision, TTQ, 
BNN, DoReFa, XNORNET}. However, a majority of existing works 
in low-precision DNNs sacrifice accuracy over
the baseline full-precision models.

Furthermore, most prior works target reducing the precision of model
parameters (network weights). This primarily benefits inference
when batch sizes are small.
We observe that activation maps (neuron outputs) occupy
more memory compared to model
parameters for typical batch sizes during training.
This observation holds even
during inference when batch size is around 16 or more.

\begin{figure}[!htb]
\begin{center}
   \includegraphics[width=1.0\linewidth]{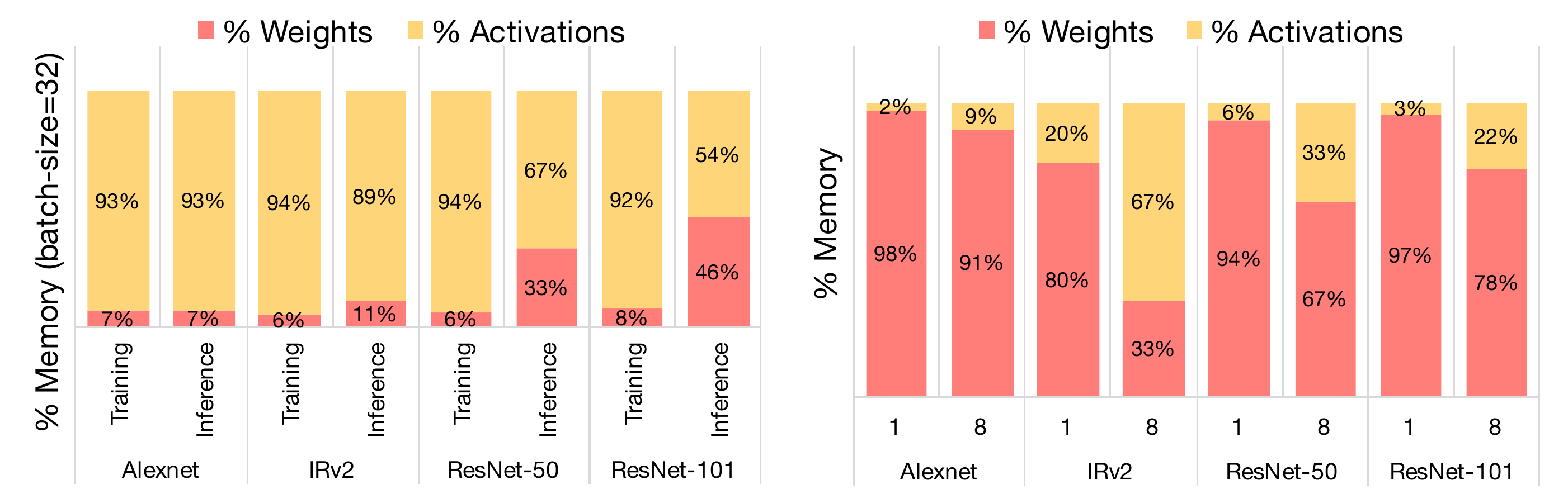}
\end{center}
   \caption{Memory footprint of weights and
   activations during training/inference (left plot), and inference with 
   batch sizes 1 and 8 (right plot).}
\label{fig:MemoryFootprint}
\end{figure}

\noindent\textbf{Motivation}: Figure~\ref{fig:MemoryFootprint} shows memory 
footprint of
activation maps and filter maps for
4 different networks -- AlexNet, Inception-Resnet-v2,
ResNet-50 and ResNet-101. The left plot in this figure shows the memory
footprint occupied by weights and activations during the training and 
inference phase with a (modestly large) batch size of 32. 
When batch-size is large, because of filter reuse across batches of
inputs, activation maps occupy significantly
larger fraction of memory compared to the filter weights.
The right plot in this figure shows the memory footprint during inference
with batch sizes of 1 and 8. With small batch size, the weights in the 
model occupy more memory compared to activations.

During training large batch sizes are typical, so lowering 
activation memory is more beneficial. It is the same case with large 
batch inference which is a common scenario for cloud based systems.
On the other hand, for real-time (low latency) inference deployments, 
batch-sizes are relatively small 
and lowering memory requirements of weights is more beneficial. 

\noindent\textbf{Our proposal}: Based on this observation, we study schemes for
training and inference using low-precision DNNs
where we reduce the precision of activation maps as well as the model
parameters without sacrificing the network accuracy.
To this end, we investigate three software/algorithm-based schemes where 
the precision of weights and activations are lowered without hurting 
network accuracy.

The next section elaborates on each of these proposals. To take 
advantage of our software-based optimizations, we also 
developed an accelerator to handle low-precision numerics. We discuss the details
of this accelerator and conclude with some discussions on next steps and trade-offs 
between low-precision and sparsity (which is another knob to lower memory 
and compute cost). 

Overall, the contributions of this paper are a few techniques to obtain 
low-precision DNNs without sacrificing model accuracy. 
Each of our schemes produces a low-precision model that surpasses
the accuracy of the equivalent low-precision model published to date. One of our schemes
also helps a low-precision model converge faster.

\section{Training and inference using low-precision}

\noindent \textbf{Wide reduced-precision networks (WRPN)~\cite{mishra2018wrpn}}: 
Based on the observation that activations occupy more memory
footprint compared to weights, we lower the precision of activations
more aggressively than weights to speed up training and inference 
steps as well as cut down on
memory requirements. However, a straightforward reduction
in precision of activation maps leads to significant reduction in
model accuracy~\cite{DoReFa, XNORNET}.

\begin{table}[!htb]
\begin{center}
   \caption{ ResNet-44 and ResNet-56 top-1 validation set error \% 
   as precision of activations (A) and weight (W)
   changes. All results are with end-to-end training
   of the network from scratch.} \label{fig:1xwide-baseline}
   \includegraphics[width=0.90\linewidth]{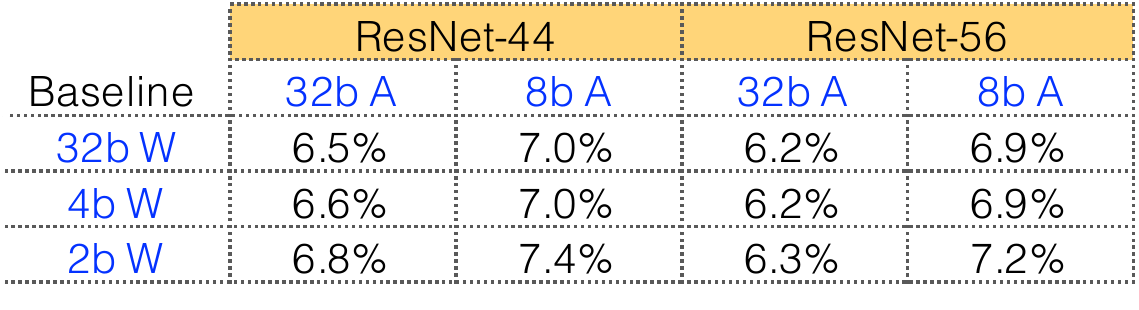}
\end{center}
\end{table}

To highlight this aspect, Table~\ref{fig:1xwide-baseline} shows a sensitivity study 
where we reduce precision of
activation maps and model weights for two ResNet topologies running CIFAR-10 
dataset and train the network from scratch.\footnote{
\noindent \textbf{ResNet topology on CIFAR-10} -- Our implementation of ResNet 
for CIFAR-10 closely follows the configuration
in~\cite{ResNet}.
The first layer is a 3$\times$3
convolutional layer
followed by a stack of 6$n$ layers with 3$\times$3 convolutions
on feature map sizes 32, 16 and 8;
with 2$n$ layers for each feature map size. The numbers of
filters are 16, 32 and 64 in each set of 2$n$ layers.}
32b weights (W) and 32b activations (A) corresponds to a baseline network
with full-precision numerics.
We find that, in general, reducing the precision of activation maps 
and weights hurts model accuracy.
Further, reducing precision of activations hurts model accuracy much more 
than reducing precision of the filter parameters. For ResNet-44, with 
2b weights and 8b activations,
there is a 0.9\% degradation in accuracy from baseline full-precision 
network (7.4\% vs. 6.5\%). For ResNet-56, this degradation is 1\% (7.2\% vs. 6.2\%).

To re-gain the model accuracy while working with reduced-precision
operands, we increase the number of filter maps in a layer and find that 
this scheme compensates or surpasses the
accuracy of the baseline full-precision network.
We call our approach \texttt{WRPN} (wide reduced-precision networks).

For both the ResNet configurations on CIFAR-10, we found that we can match accuracy of 
a 2b weight and 8b activation model by \textit{doubling} the number of 
filters of the entire network.

Table~\ref{fig:1_3xwide-wrpn} shows the effect on accuracy as precision is lowered
and filters widened by a smaller factor than 2x -- only the filters in the 
first 1/3rd of the layers are doubled, the remaining layers have the same number of filters
as baseline network. With this setting, the accuracy with 2b weights
and 8b activations is within 0.5\% of baseline full-precision accuracy. 
Using \texttt{WRPN} with a technique that we will describe next, the accuracy of the network
at this precision knob matches with that of baseline. 

\begin{table}[!htb]
\begin{center}
    \caption{ ResNet-44 and ResNet-56 top-1 validation set error \% 
      as precision of activations (A) and weight (W)
      changes. The number of filters in the first 30\% of the layers are doubled. 
      All results are with end-to-end training
      of the network from scratch. }
   \label{fig:1_3xwide-wrpn}
   \includegraphics[width=0.90\linewidth]{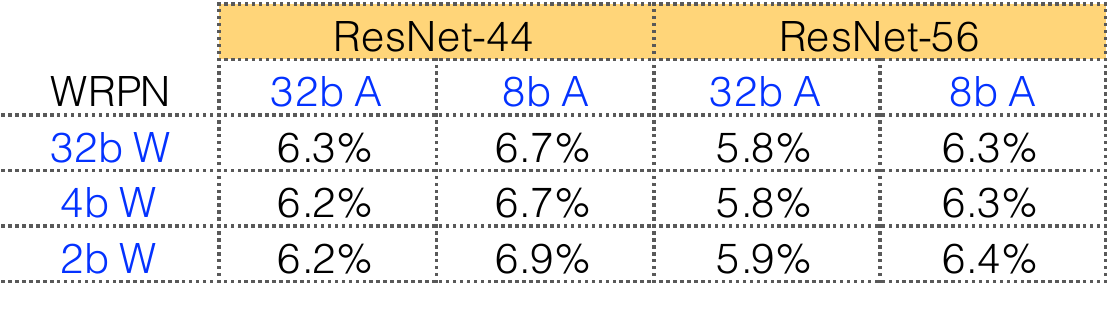}
\end{center}
\end{table}

In practice, we find \texttt{WRPN} scheme to be very simple and effective
- starting with a baseline network architecture, one can change the
width of each filter map without changing any other
network design parameter or hyper-parameters.
Reducing precision and simultaneously
widening filters keeps the total compute cost of the network
under or at-par with baseline cost.\footnote{Compute cost is the 
product of the number of FMA operations
and the sum of width of the activation and weight operands.}
Thus, although the number of raw compute operations increase
with widening the filter maps in a layer,
the bits required per compute operation is now a fraction of
what is required when using full-precision operations.
As a result, with appropriate hardware support, one can significantly
reduce the dynamic memory requirements, memory bandwidth,
computational energy and speed up the training and inference process.

\noindent \textbf{Apprentice~\cite{mishra2018apprentice}}: With this scheme we combine network quantization schemes
with model compression techniques
and show that the accuracies of low-precision networks can be significantly improved by
using knowledge distillation techniques.
Previous studies on
model compression use a large network as the teacher network
and a small network as the student network. The small student network learns from the
teacher network using a technique called knowledge distillation~\cite{Hinton-distill, Compression-Caruana, 2016arXiv160305691U}.

The network architecture of the student network
is typically different from that of the teacher network --
for e.g.~\cite{Hinton-distill} investigate a student
network that has fewer number of neurons in the hidden layers compared to the
teacher network. In our work, the student network has similar
topology as that of the teacher network, except that the
student network has low-precision
neurons compared to the teacher network which has neurons
operating at full-precision. 

\begin{figure}[!htb]
\begin{center}
   \includegraphics[width=0.85\linewidth]{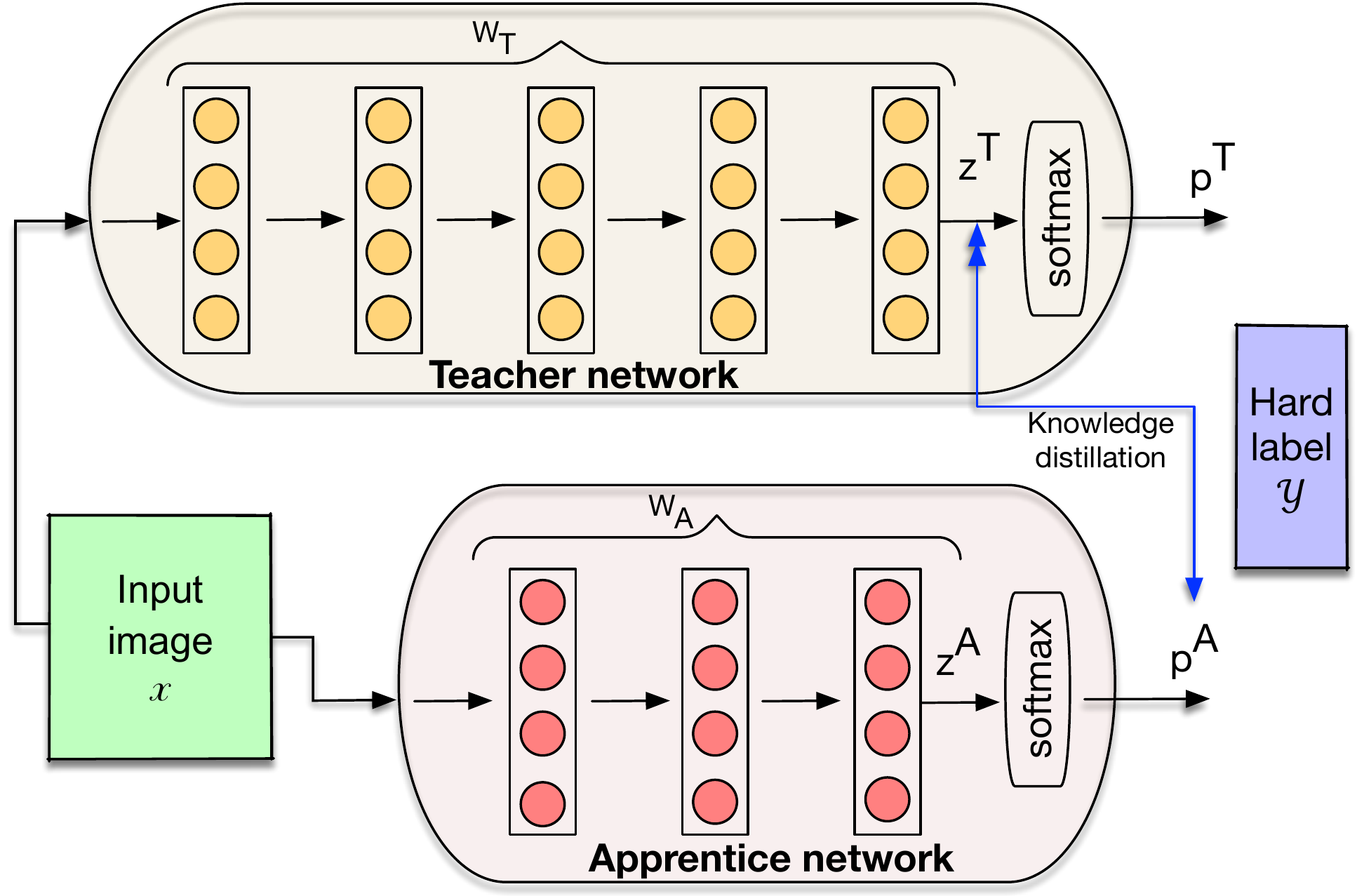}
\end{center}
   \caption{ Schematic of the knowledge distillation setup.}
\label{fig:apprentice}
\end{figure}

We call our approach \texttt{Apprentice} where
a low-precision (student/apprentice) network is learning (and attempting to mimic) 
the knowledge of a 
high precision (teacher) network. With this scheme, we start with a 
full-precision trained (teacher) network and transfer knowledge from
this trained network continuously to train a low-precision (student) 
network from scratch. The knowledge transfer process consists of matching 
the logits of the teacher network with the softmax scores of the student network.
This is included as an addition term in the cost function while training. 
Figure~\ref{fig:apprentice} shows a schematic of the training setup.
We find that the low-precision network converges faster and to better accuracy (compared 
to the student network being trained alone) when a trained complex network 
guides its training.

\begin{table}[!htb]
\begin{center}
       \caption{ ResNet-44 and ResNet-56 top-1 validation set error \% 
       as precision of activations (A) and weight (W)
       changes. The networks are trained under the supervision of a full-precision 
       ResNet-101 network using distillation based technique.}
    \label{fig:1xwide-apprentice}
   \includegraphics[width=0.90\linewidth]{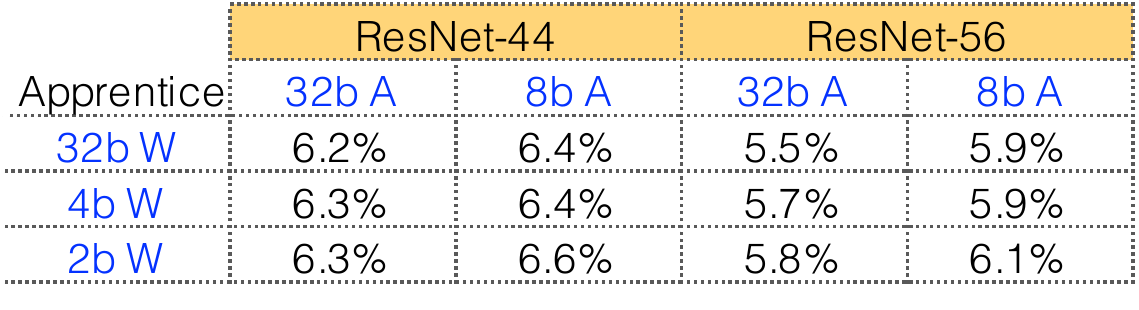}
\end{center} 
\end{table}

Table~\ref{fig:1xwide-apprentice} shows the impact of lowering precision
when a low-precision (student) network is paired with a full-precision
(ResNet-101) network. \textit{We find the Apprentice scheme to improve the 
baseline full-precision
accuracy}. The scheme also helps close the gap between the new improved
baseline accuracy and the accuracy when lowering
the precision of the weights and activations.
The gap between 2b weight and 8b activation ResNet-44 is
now 0.2\%. For ResNet-56, this gap is 0.6\%. 

\begin{table}[!htb]
\begin{center}
    \caption{ ResNet-44 and ResNet-56 top-1 validation set error \% 
    as precision of activations (A) and weight (W)
    changes. The networks are trained under the supervision of a full-precision 
    ResNet-101 network using distillation based technique. 
    The filters in the first 30\% of the layers are doubled. }
    \label{fig:wrpn-apprentice}
   \includegraphics[width=0.90\linewidth]{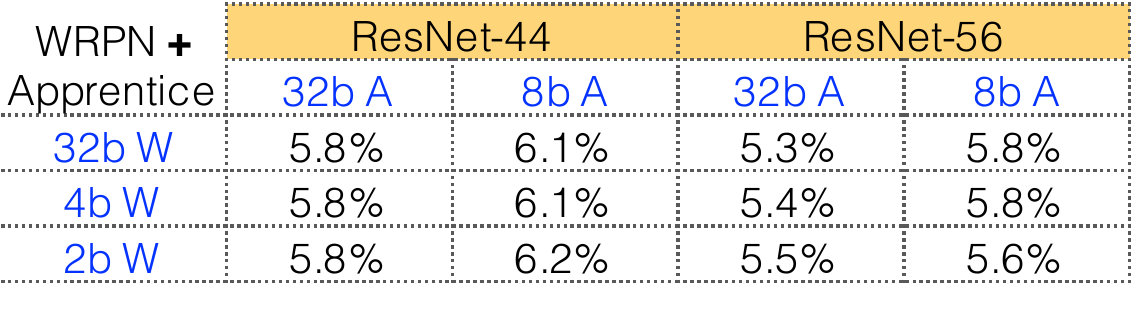}
\end{center}
\end{table}

\noindent \textbf{WRPN+Apprentice}: This scheme combines widening of filters with 
distillation scheme. The training process is the same as the \texttt{Apprentice} 
scheme. Table~\ref{fig:wrpn-apprentice} shows the results with this scheme on ResNet.
With 2b weights and 8b activations, ResNet-44 is 0.3\% better than the 
baseline we started with (6.2\% now vs. 6.5\% baseline). Similarly, ResNet-56 is 
0.6\% better than the baseline (5.6\% now vs. 6.2\% baseline).

Overall, each of the three schemes described above improve the
accuracy of the low-precision network configuration compared to baseline as well as
prior proposals. 

\begin{figure}[!htb]
\begin{center}
   \includegraphics[width=0.80\linewidth]{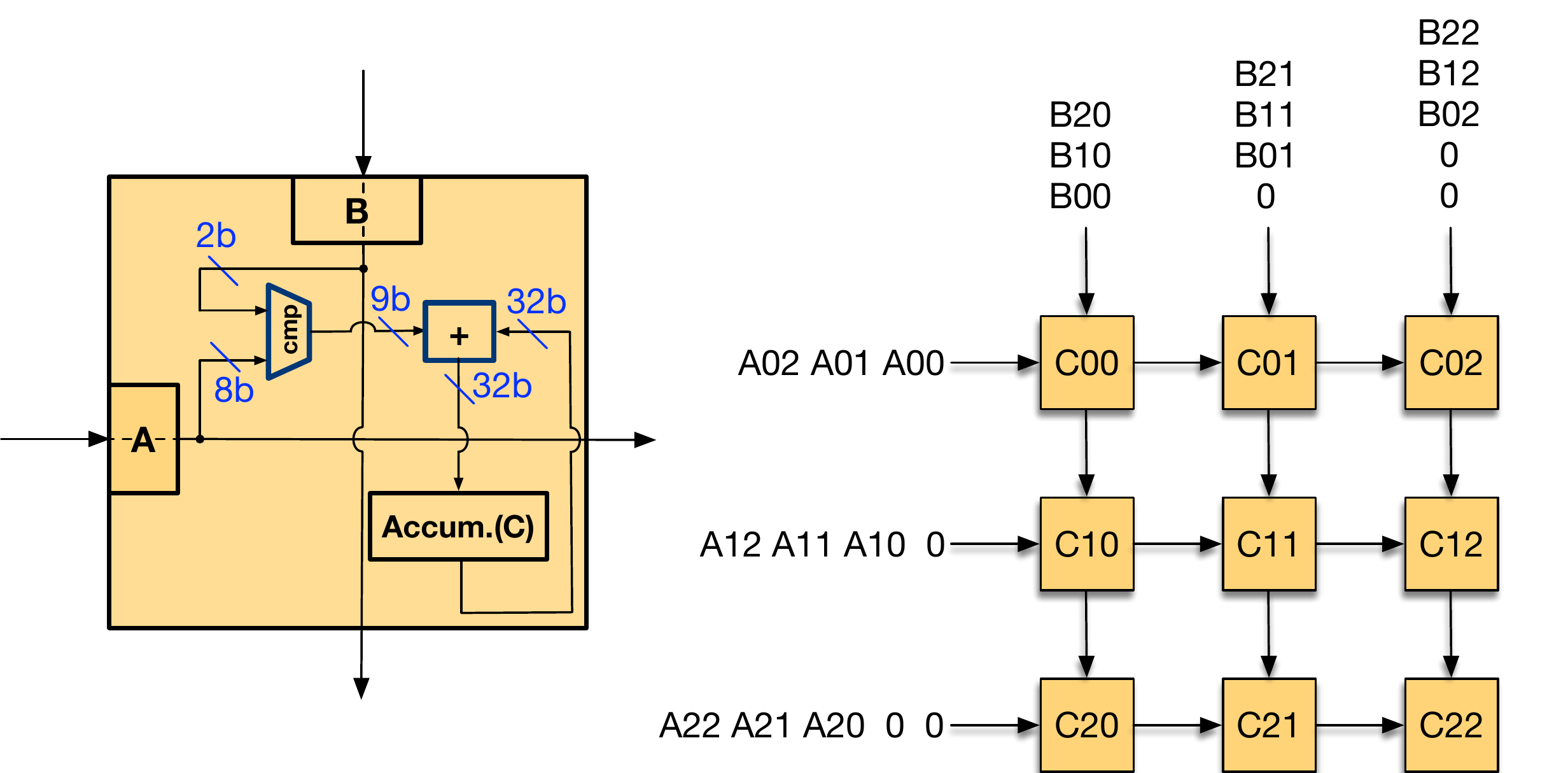}
\end{center}
   \caption{ Accelerator for 8b activations and 2b weights.}
\label{fig:accel}
\end{figure}

\section{Low-precision GEMM accelerator} \label{sec:accelerator}
To take advantage of the low-precision numerics, we developed an accelerator
optimized for 2b$\mathrm{x}$8b matrix multiplication (convolution operation can be mapped
as a series of small GEMMs). Figure~\ref{fig:accel} shows the schematic of 
the accelerator (organized as a systolic array) and the processing engine in each 
of the systolic cross-points. 
The accelerator
consists of 64 processing engines organized in 8 rows and 8 columns.
In the figure, $A$ matrix corresponds to activation tensor and consists of unsigned integer values (since
post-ReLU only positive values are significant). 
With 2b operand, we do not need a 
floating-point or integer multiplier. With 2b weights 
(corresponding to $B$ matrix in the figure), 
the possible weight values are \{-1, 0, 1\}. Thus, to multiply an 8b integer ($A$) 
with a weight value ($B$), we compare the sign of $B$ operand and append this to the
$A$ operand or make it zero. The accumulator is a 32b signed integer.  
Compared to a full-precision accelerator, 
at the same technology process node, our low-precision accelerator is 
15$\mathrm{x}$ smaller in area and 12$\mathrm{x}$ efficient in power.

\section{Conclusions} \label{sec:conclusions}

With vendors packing more and more compute (FLOPs) on a die, bandwidth is becoming 
a key issue that affects scaling in large-scale systems. Low-precision 
techniques help
in these scenarios.
While low-precision networks have system-level benefits, the 
drawback of such models is degraded accuracy when compared to  
full-precision models.
We present schemes to improve the accuracy
of low-precision networks and close the gap between the accuracy of
these models and full-precision models. Each of the schemes
improve the accuracy of the low-precision
network configuration compared to prior proposals. We motivate the 
need for a smaller model size
in low batch, real-time and resource constrained inference deployment systems. 
For large batch mode, we discuss benefits of lowering the 
precision of activation maps as opposed 
filter weights. Finally, we discuss the 
design of an accelerator optimized for ternary-precision.

\bibliographystyle{IEEEtranS}
\renewcommand{\baselinestretch}{0.20} 
\bibliography{egbib}

\end{document}